\documentclass{article}

\usepackage[preprint]{neurips_data_2023}
\usepackage{gensymb}
\usepackage[utf8]{inputenc} % allow utf-8 input
\usepackage[T1]{fontenc}    % use 8-bit T1 fonts
\usepackage{hyperref}       % hyperlinks
\usepackage{url}            % simple URL typesetting
\usepackage{booktabs}       % professional-quality tables
\usepackage{amsfonts}       % blackboard math symbols
\usepackage{nicefrac}       % compact symbols for 1/2, etc.
\usepackage{microtype}      % microtypography
\usepackage{xcolor}         % colors
\usepackage{graphicx}
\usepackage{enumitem}
\usepackage{amsmath}
\usepackage{chngcntr}
\usepackage{dirtree}
\usepackage{listings}
\setlist{nosep}
\usepackage{hyperref}
\hypersetup{
    colorlinks=true,
    linkcolor=blue,
    filecolor=magenta,      
    urlcolor=cyan,
    pdftitle={Overleaf Example},
    pdfpagemode=FullScreen,
    }

\definecolor{codegreen}{rgb}{0,0.6,0}
\definecolor{codegray}{rgb}{0.5,0.5,0.5}
\definecolor{codepurple}{rgb}{0.58,0,0.82}
\definecolor{backcolour}{rgb}{0.95,0.95,0.92}

\lstdefinestyle{mystyle}{
    backgroundcolor=\color{backcolour},   
    commentstyle=\color{codegreen},
    keywordstyle=\color{magenta},
    numberstyle=\tiny\color{codegray},
    stringstyle=\color{codepurple},
    basicstyle=\ttfamily\footnotesize,
    breakatwhitespace=false,         
    breaklines=true,                 
    captionpos=b,                    
    keepspaces=true,                 
    numbers=left,                    
    numbersep=5pt,                  
    showspaces=false,                
    showstringspaces=false,
    showtabs=false,                  
    tabsize=2
}

\usepackage{chngcntr}
\setlist{nosep}

\title{Verification against in-situ observations for \\ Data-Driven Weather Prediction}

\author{%
  Vivek Ramavajjala \\
  Excarta Inc.\\
  \texttt{vivek@excarta.io} \\
   \And
   Peetak P. Mitra \\
   Excarta Inc.\\
   \texttt{peetak@excarta.io} \\
}

\begin{document}

\maketitle

\begin{abstract}
Data-driven weather prediction models (DDWPs) have made rapid strides in recent years, demonstrating an ability to perform medium-range weather forecasting a high degree of accuracy. The fast, accurate, and low-cost forecasts promised by DDWPs make their use in operational forecasting an attractive proposition, however, there remains work to be done in rigorously evaluating DDWPs in a true operational setting. DDWPs have been trained and evaluated only using the ERA5 reanalysis dataset, which is produced using a combination of real-world weather observations and a Numerical Weather Prediction (NWP) model. While the ERA5 dataset represents a very high quality simulation of global weather, it is still influenced by the underlying NWP and is not a replacement for real-world observations. Additional benchmarks are needed to examine how well DDWPs perform when deployed in operational settings and tested against real-world observations. Such benchmarks can provide unbiased measures of model quality, and build confidence in DDWPs as they are deployed in operational settings. We present a robust dataset of in-situ observations derived from the NOAA Meteorological Assimilation Data Ingest System (MADIS) program to serve as a benchmark to validate DDWPs in an operational setting. We use this dataset to compare a well-known DDWP (FourCastNet) against an NWP baseline (IFS), and observe that even though FourCastNet outperforms IFS when tested against ERA5, it does not have a performance benefit when tested against real-world observations. By providing a large corpus of quality-controlled, in-situ observations, this dataset provides a meaningful real-world task that all NWPs and DDWPs can be validated against. We hope that this data can be used not only to rigorously and fairly compare operational weather models but also to spur future research in new directions.
\end{abstract}

\section{Introduction}
%\vspace{-1mm}
Data-driven weather prediction (DDWP) models have made rapid strides in recent years, steadily improving forecast skill over longer time horizons at a fraction of the cost of Numerical Weather Prediction (NWP) models. Given the demonstrated improvements in forecast skill, low inference cost, and the ability to target specific variables of interest, using DDWPs in operational weather forecasting is an attractive proposition. Weather forecasts play a foundational role in day-to-day operations of key sectors like agriculture, energy, transportation, and in reducing loss of life due to extreme weather events, thus allowing DDWPs to have a tangible social benefit. This potential impact also makes it imperative that ML practitioners should take a holistic view of DDWPs in the \textit{operational} forecasting setting, testing DDWPs under the same constraints faced by NWPs that currently underpin operational forecasts. Similar examination of ML models is common in the development of large language models (LLMs), where a "red-teaming" approach is used to identify how a robustly trained LLM may yet produce undesirable outcomes in a real-world scenario (e.g., \cite{perez2022red}, \cite{tamkin2021understanding}). For DDWPs, a good starting point is defining evaluation tasks that more accurately reflect their use in operational forecasts.\par

The most promising DDWPs developed recently have been trained using the ERA5 reanalysis dataset (\cite{hersbach2020era5}), typically using data from 1979-2017 for training and development, and using data from 2018 as a held-out test set. Produced by combining historical observations and a NWP, the ERA5 data provides "maps without gaps" of essential climate variables (ECVs). The ML models use a fixed subset of ECVs from ERA5 to represent the atmosphere at time \textit{t}, and predict the same subset of ECVs at \textit{t+$\delta$}, where $\delta$ is most commonly 6 hours, but varies from 1 hour to 24 hours. All ECVs are typically normalized to zero mean and unit variance, and a mean-squared error (MSE) loss is used to train the models. For evaluation, DDWPs are typically initialized with data from ERA5 at 00UTC on a particular day, and iteratively rolled out to generate a forecast for the next 14 days. The forecasts are evaluated by comparing the predictions against ERA5 data using two metrics: the root mean squared error (RMSE) and the anomaly correlation coefficient (ACC), computed over the entire latitude-longitude grid, and weighted in proportion to the latitude. To establish a NWP baseline, the RMSE and ACC metrics are similarly calculated for ECMWF's Integrated Forecasting System (IFS). For details, we refer to FourCastNet (\cite{pathak2022fourcastnet}), Pangu-Weather (\cite{bi2022pangu}), or GraphCast (\cite{lam2022graphcast}), which share essentially the same procedure for training and evaluating different DDWPs. DDWPs thus learn a high-quality approximation of the underlying NWP used to construct the ERA5 dataset, and evaluations using ERA5 data test the quality of the learned approximation. \par
%%\vspace{-0.5mm}
\subsection{Limitations of verification against reanalysis}
%%\vspace{-0.5mm}
Verifying DDWP forecasts against ERA5 reanalysis is similar to the "verification against analysis" procedure used to evaluate operational NWPs, where the forecast from an NWP is compared to a sequence of analyses produced from the same NWP\footnote{For instance, see "grid-to-grid" verification against analysis published by NOAA: \url{https://www.emc.ncep.noaa.gov/users/verification/global/gfs/ops/grid2grid_all_models/rmse/}.}. While convenient, since the sequence of analyses depends on the NWP output from the previous cycle, such evaluations can underestimate the forecast error, especially for short lead times (\cite{bowler2015verification}, \cite{bowler2008accounting}). Besides fundamental limitations of verification against analysis, comparing against ERA5 is not sufficient to estimate how well the DDWP performs in an operational setting, both objectively and relative to the IFS baseline.\par
First, the analysis available for initializing operational DDWPs assimilate observations from +3 hours ahead, and may not be as robust as the ERA5 reanalysis, which assimilates observations from +9 hours at 00UTC. Second, evaluating IFS on the ERA5 dataset is not necessarily meaningful since it has not been optimized to predict ERA5 data, whereas DDWPs have --- giving DDWPs an unfair advantage. Further, since the ERA5 dataset uses an underlying NWP to create a picture of the global weather, it is inherently biased towards the underlying NWP and its assumptions, and hence may differ from real weather conditions, creating a noticeable difference between simulation and reality. Indeed, simply because a DDWP outperforms the IFS model when tested against ERA5 does not guarantee the same performance improvement holds in an operational setting, where both start states and ground truth are different. Such concerns have already been raised by \cite{bi2023accurate}, who note the limitations of using only reanalysis data for validation despite real-world systems working with observational data. We lastly note that operational NWPs are verified not just against analysis, but also verified against observations, e.g. ECMWF scorecards report performance against both analysis and observations\footnote{See \url{https://sites.ecmwf.int/ifs/scorecards/scorecards-48r1ENS.html}}. This underscores the need for similar evaluation benchmarks for DDWPs.

In domains such as robotics or RL-based control, the "sim2real" challenge of transferring and evaluating a model trained via simulation in a real-world setting is well known (c.f., \cite{kadian2020sim2real}). Viewing the ERA5 dataset as a high-quality simulation of weather, there is a similar need for a "sim2real" step that evaluates DDWPs against observations. We also note that both the industry and common user primarily consume weather forecasts for specific locations (vs.  gridded forecasts), making such "verification against observations" a meaningful test of operational weather models.\par
%\vspace{-0.5mm}
\subsection{Our contributions}
%\vspace{-0.5mm}
\label{ourcont}
We introduce a dataset of quality-controlled in-situ observations derived from the NOAA Meteorological Assimilation Data Ingest System (MADIS) program that can be used to more thoroughly evaluate the performance of DDWPs in an operational setting. Additionally, since in-situ observations are not modeled in any way but instead reflect the actual weather on the ground, they can be used for unbiased comparisons of DDWPs and NWPs on a task meaningful to consumers of forecasts. In Section~\ref{verifsection}, we compare FourCastNet DDWP model (\cite{pathak2022fourcastnet}) and the IFS model against real-world observations. FourCastNet has previously shown improvements over the IFS model when evaluated against ERA5, here we examine its performance against real-world observations.\par

Prior work has examined ERA5 data against observations for specific variables and regions, e.g., \cite{jiao2021evaluation} evaluate ERA5 for precipitation in China, or to assess its suitability for specific tasks, e.g., choosing suitable location for wind farms (\cite{olauson2018era5}. To our knowledge, this is the first publicly available comparison of ERA5 against in-situ observations at such a scale, as well as the first publicly available dataset for verifying NWPs and DDWPs against observations. The dataset is made available at \url{https://huggingface.co/datasets/excarta/madis2020}.

\section{Dataset}
%\vspace{-0.5mm}
\label{dataset}
\subsection{Background of the MADIS database}
%\vspace{-0.5mm}
MADIS\footnote{\url{https://madis.ncep.noaa.gov/index.shtml}} (Meteorological Assimilation Data Ingest System) is a database provided by NOAA (National Oceanic and Atmospheric Administration) that contains meteorological observations covering the entire globe. MADIS ingests data from NOAA and non-NOAA sources, including observation networks ("mesonets") from US federal, state, and transportation agencies, universities, volunteer networks, and data from private sectors like airlines as well as public-private partnerships like CWOP (Citizen Weather Observer Program). Whereas reanalysis data is influenced by the underlying physics of the NWP used to reconstruct the reanlaysis, these observations directly capture the weather on the ground and can be used to objectively evaluate any data-driven or numerical weather model, without being biased towards either of the approaches. 
%\vspace{-0.5mm}
\subsection{Curating the MADIS2020 dataset}
%\vspace{-0.5mm}
We created a dataset "MADIS2020" consisting of hourly observations from the Mesonet and METAR networks ingested by MADIS\footnote{Not all sources ingested by MADIS are free for public use, we choose Mesonet and METAR sources as they are publicly usable, and have large data volumes.} for the entirety of 2020, totaling over 700 million quality-checked observations. Mesonet observations represent data from different weather observation networks, while METAR data primarily indicates data collected at airports. We have additionally taken care to use the subset of MADIS data fully open to public\footnote{\url{https://madis.ncep.noaa.gov/mesonet_providers.shtml}}, and to follow the FAIR principles \cite{wilkinson2016fair} in aggregating and disseminating the data. Neither the raw nor the curated data contains any personally identifiable information.

%\vspace{-0.5mm}
\subsection{Dataset features and the quality thresholds}
%\vspace{-0.5mm}
We limit ourselves to a few key surface meteorological variables that have an outsized impact on society, discussed in Table \ref{var-table} :

\begin{table}
  \caption{Variables in dataset, provided at an hourly frequency for the entire year of 2020}
  \label{var-table}
  \centering
  \begin{tabular}{llll}
    \toprule
    Variable name    & Description &  Short name  & Units \\
    \midrule
    Temperature  & Air temperature at \texttt{2m} height      & \texttt{t2m}   & $\mathtt{K}$  \\
    Dewpoint     & Dewpoint temperature at \texttt{2m} height & \texttt{d2m}   & $\mathtt{K}$       \\
    Wind speed   & Windspeeds calculated from \texttt{10m} \texttt{u} and \texttt{v} components   & \texttt{wind}   & $\mathtt{ms^{-1}}$  \\
    \bottomrule
  \end{tabular}
\end{table}

Temperature, dewpoint, and wind speeds are instantaneous observations and not averaged over a time window. The MADIS database also includes quality control flags that specify the level to which an observation has been checked\footnote{\url{https://madis.ncep.noaa.gov/madis_sfc_qc_notes.shtml}}:
\begin{itemize}
    \item Level 1: Observation is checked to lie within a feasible range:
        \subitem Temperature between $\mathtt{-60F}$ to $\mathtt{130F}$
        \subitem Dewpoint between $\mathtt{-90F}$ to $\mathtt{90F}$
        \subitem Wind speed must lie between $\mathtt{0kts}$ - $\mathtt{250kts}$
    \item Level 2: Internal consistency, temporal consistency, and statistical spatial consistency are checked. Internal consistency checks enforce meteorological relationships between different variables at the same station, e.g., dewpoint must not exceed temperature. The temporal consistency check flag observations that change too rapidly, and the statistical spatial consistency check flag observations that have failed quality checks 75\% of the time in the last 7 days.
    \item Level 3: Spatial consistency checks that verify one observation against nearby observations.
\end{itemize}

Based on the QC levels above, different quality flags are defined and applied for each observation:
\begin{itemize}
    \item \texttt{C}: Coarse pass, passed level 1
    \item \texttt{S}: Screened, passed levels 1 and 2
    \item \texttt{V}: Verified, passed levels 1, 2, and 3
    \item \texttt{X}: Rejected, failed level 1
    \item \texttt{Q}: Erroneous, passed level 1 but failed level 2 or 3
\end{itemize}

We include only observations that have QC flags \texttt{S} or \texttt{V}, i.e., passing at least level 2 checks. Some observations pass level 2 checks, but are in regions too data-sparse to support level 3 checks,  and may still introduce noisy observations. To remove such noisy observations, we ignore temperature and dewpoint observations that were more than 20 C away from the ERA5 prediction. 

\subsection{Dataset geographical coverage}

We partition the observations into different geographical regions, extending from the list of regions used in ECMWF scorecards\footnote{\url{https://sites.ecmwf.int/ifs/scorecards/scorecards-48r1ENS.html}}, graphically shown in the Appendix figure \ref{figa1} and the latitude/longitude extents are provided in Table \ref{geo-table}.

\begin{figure}[!htb]
    \centering
    \includegraphics[scale=0.4]{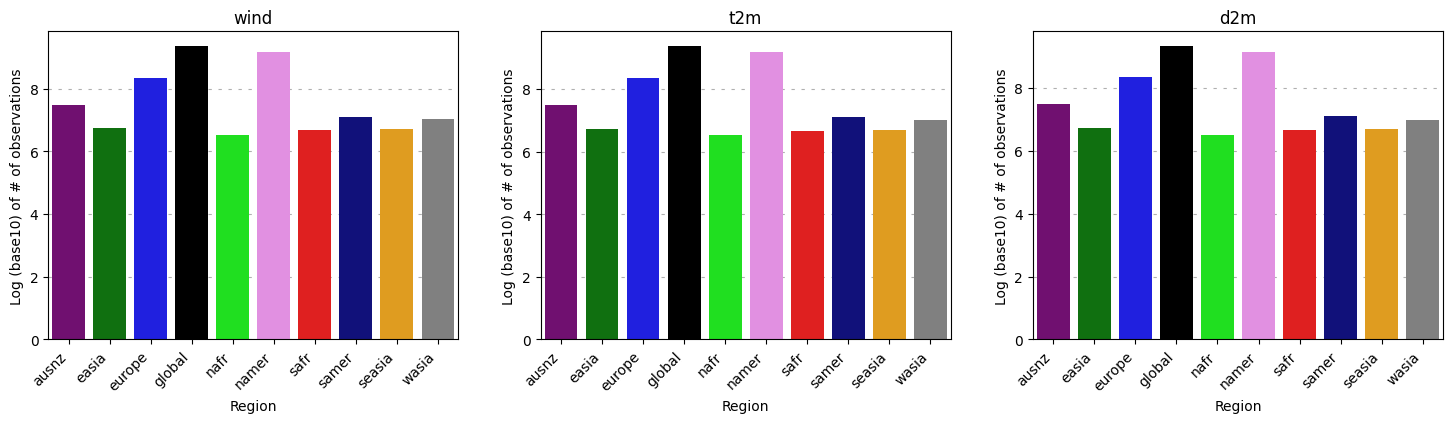}
    \caption{Total number of observations for different variables in 2020, in log (base 10) scale.}
    \label{fig:obscount}
\end{figure}

It is to be noted that the above regions are not mutually exclusive, and a small number of observations in the "global" region do not fall under any of the defined regions. Figure \ref{fig:obscount} shows the distribution of observations globally and for each region. All regions have at least 1,000,000 observations for temperature, wind speeds, and dewpoint. Unsurprisingly, these data are not uniformly distributed across the world, with North America and Europe having the greatest density of observations, and the global South having a much lower density of observation data.

\subsection{Comparing observations against ERA5}
\begin{figure}
    \centering
    \includegraphics[scale=0.4]{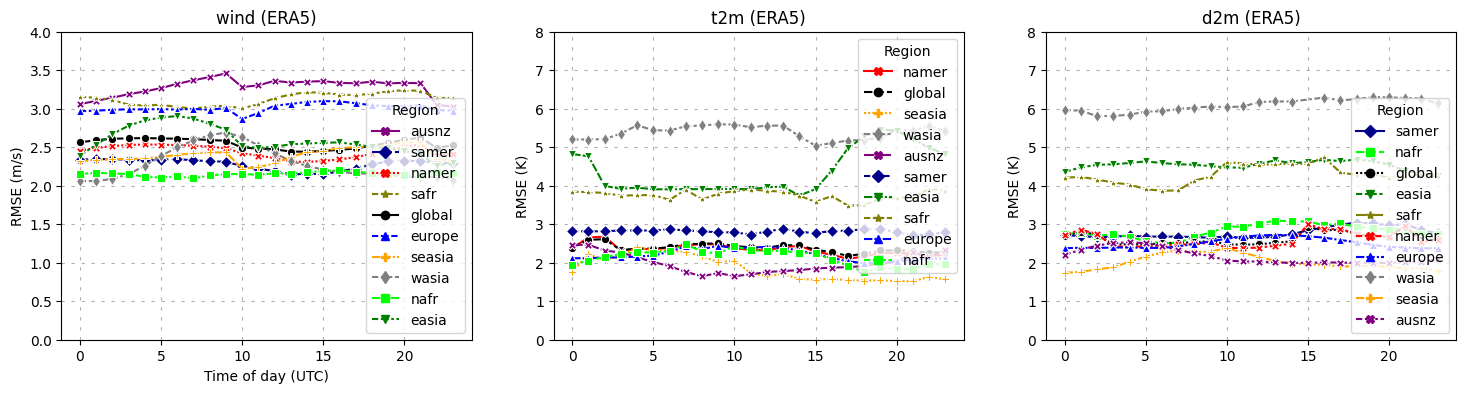}
    \caption{RMSE of ERA5 data vs. real-world observations for different regions.}
    \label{fig:madisera5}
\end{figure}

We first test how well the ERA5 dataset corresponds to real-world observations, we compare each hourly observation for all weather stations in MADIS2020 with the corresponding ERA5 estimate, using RMSE to quantify the error for each region. To identify diurnal variation in quality, we report the RMSE of ERA5 vs. observations for separately for each hour of the day. Figure ~\ref{fig:madisera5} shows the RMSE for wind speeds, temperature, and dewpoint. We note that ERA5 shows a consistent performance across all regions for wind speeds, but has a notable regional variation in quality for temperature and dewpoint. In particular, regions in Asia and the global South have higher RMSE for temperature and dewpoint.\par
Since this regional variation exists in the data itself, it is expected that any models trained on ERA5 data would have similar regional variations as well, and should be kept in mind when training models on ERA5 data.
%\vspace{-2mm}
\section{Validating a Data-Driven Weather Prediction model}
%\vspace{-1mm}
\label{verifsection}
Having set up the MADIS2020 dataset, we use it to examine whether a DDWP performs as well against real-world data as it does in simulations. We choose FourCastNet as the DDWP to evaluate, since it has a public implementation and is relativel easy to replicate. As with prior work in DDWPs, we use the IFS forecast from ECMWF as an NWP baseline.

\subsection{Training the DDWP}
We train a variant of the FourCastNet (FCN) model, using the same model architecture and hyperparameters described by \cite{pathak2022fourcastnet}. The original FCN implementation\footnote{A public implementation is available at \url{https://github.com/NVlabs/FourCastNet}.} produces 6-hourly forecasts, and we train an additional FCN model to make 1-hourly forecasts, using 1-hourly ERA5 data but keeping other training and modeling choices the same as the 6-hourly model. We use a hierarchical temporal aggregation process (similar to the process followed by \cite{bi2022pangu}) to interleave the two models and obtain hourly forecasts at 0.25\textdegree{} resolution. Additionally, we incorporate dewpoint at 2m height as an additional input and output variable.

\subsection{Verification against reanalysis}
Since MADIS2020 contains observations only from 2020, we use ERA5 data from 2020 to validate FCN and IFS. Following the prior AI-based methods, we initialize the FCN model using ERA5 data at 00UTC and produce 48-hour forecasts. We compute the latitude-weighted RMSE and Anomaly Correlation Coefficient (ACC) metrics as defined below:
\begin{align*}
    \text{RMSE}(v,t) &= \sqrt{\frac{\sum_{i=1}^{N_{lat}}\sum_{j=1}^{N_{lon}}L(i)(\hat{X}^{v}_{i,j,t}-X^{v}_{i,j,t})^2}{N_{lat}N_{lon}}} \\
    \text{ACC}(v,t) &= \frac{\sum_{i,j}L(i)\hat{X}'^{v}_{i,j,t}X'^{v}_{i,j,t}}{
    \sqrt{\sum_{i,j}L(i)(\hat{X}'^{v}_{i,j,t})^2\sum_{i,j}L(i)(X'^{v}_{i,j,t})^2}
    } 
\end{align*}
where $L(i) = N_{\text{lat}}\frac{\cos{\phi}}{\sum_{i'=1}^{N_{\text{lat}}}\cos{\phi_{i'}}}$ stands for the weight at latitude $\phi_{i}$ and \textbf{X'} denotes the difference between $X$ and the climatology. We calcuate the RMSE and ACC metrics for temperature, dewpoint, and u10 (the eastward component of windspeed). Following prior work in DDWPs, We exclude v10, the northward component of windspeed, as it is highly correlated with u10.  A lower RMSE and higher ACC indicate a better performing model, we refer to \cite{pathak2022fourcastnet}, \cite{bi2022pangu}, \cite{lam2022graphcast} for more detailed comparisons on other DDWPs againste ERA5.\par
\begin{figure}[!htb]
    \centering
    \includegraphics[scale=0.4]{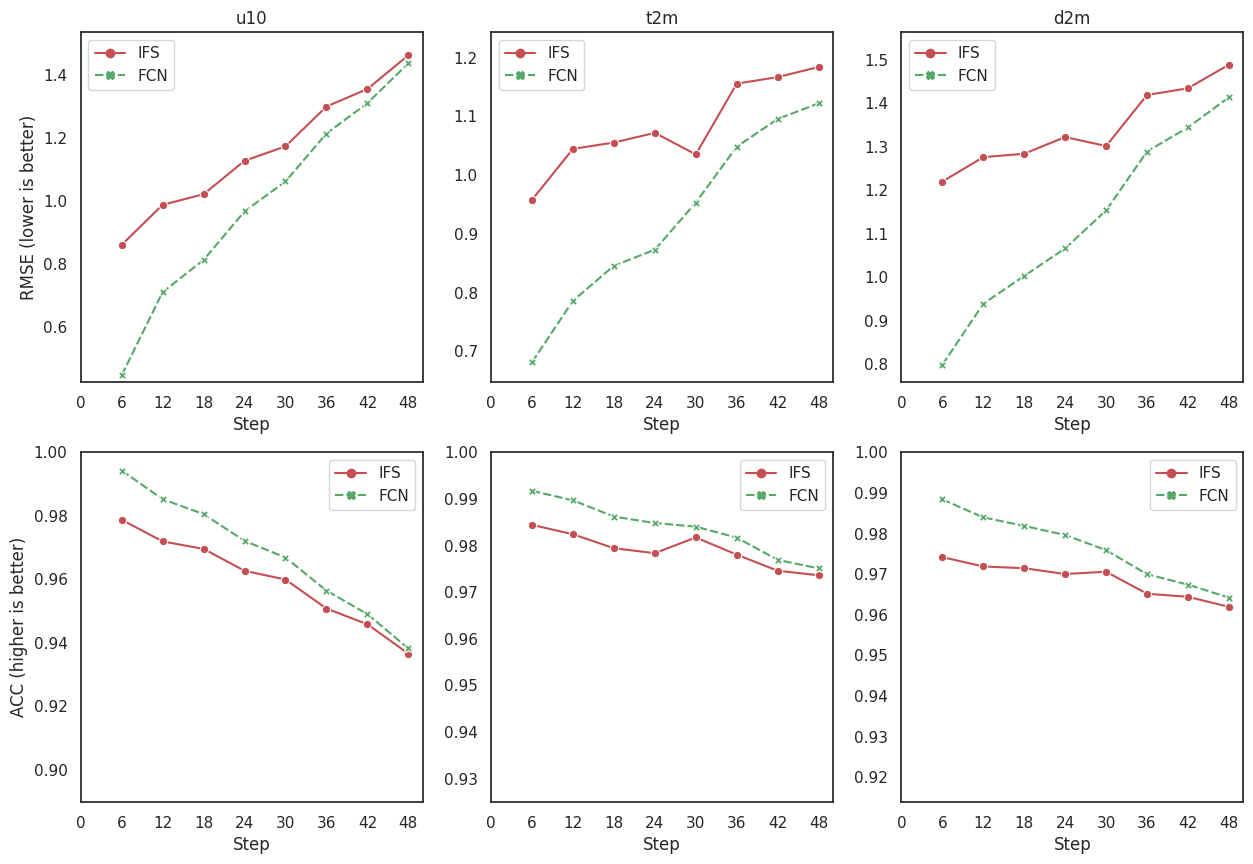}
    \caption{RMSE and ACC comparison for FourCastNet (FCN) and IFS for a 48 hour forecast. The top row shows RMSE, while the bottom row shows the ACC for each model and variable. In all cases, the FCN model outperforms IFS.}
    \label{fig:era5valid}
\end{figure}

Figure ~\ref{fig:era5valid} shows the RMSE and ACC for FCN and IFS for the first 48 hours of forecast lead time. Compared to IFS, FourCastNet clearly shows a reduction in RMSE and improvement in ACC, validating previously published results for FourCastNet.

\subsection{Verification against observations}
\subsubsection{Initializing the DDWP}
To validate the FCN model in a real-world setting, it needs to be initialized using operational analysis data and not reanalysis data from ERA5. The ideal test would be to initialize FCN with the same analysis used to initialize IFS, and while such data is available from the TIGGE archive, very long download times made this impracticable. To approximate the lower quality of an operational start state vs. a reanalysis start state, we construct start states using the \textit{ensemble mean} data from ERA5 instead of the \textit{HRES} data from ERA5. The ensemble mean data in ERA5 is produced at a lower resolution (63km) than the HRES data (31km)\footnote{\url{https://confluence.ecmwf.int/display/CKB/ERA5\%3A+data+documentation\#heading-Spatialgrid}}, thus making it a coarser reanalysis compared to HRES. In our experiments, we saw a noticeable drop in the quality of predictions made using ensemble mean start states compared to HRES start states, mimicking the effect of having lower quality start states in operational settings.\par
\subsubsection{Evaluating gridded forecasts on point observations}
FCN and IFS forecasts are produced at a 0.25\textdegree{} grid resolution, whereas real-world observations in MADIS2020 are at specific locations that may not coincide with grid points. We use bilinear interpolation to obtain forecasts at specific points from gridded forecasts, using the functionality provided by the Xarray software package by \cite{hoyer2017xarray}. While bilinear interpolation is not a perfect way of interpolating a gridded forecast to a specific location, we note that the same limitations apply to FCN and IFS as they are both produced on identical grids. Hence, bilinear interpolation provides an unbiased way to evaluate both DDWPs and NWPs on point observation data. We note that any improved interpolation methods, e.g., ones that account for latitude distortion in the grid, would equally benefit NWPs and DDWPs. Once gridded forecasts are converted to point forecasts, we can compute the RMSE for each variable (wind speed, temperature, dewpoint) for each forecast and region.\par

\begin{figure}[!htb]
    \centering
    \includegraphics[scale=0.4]{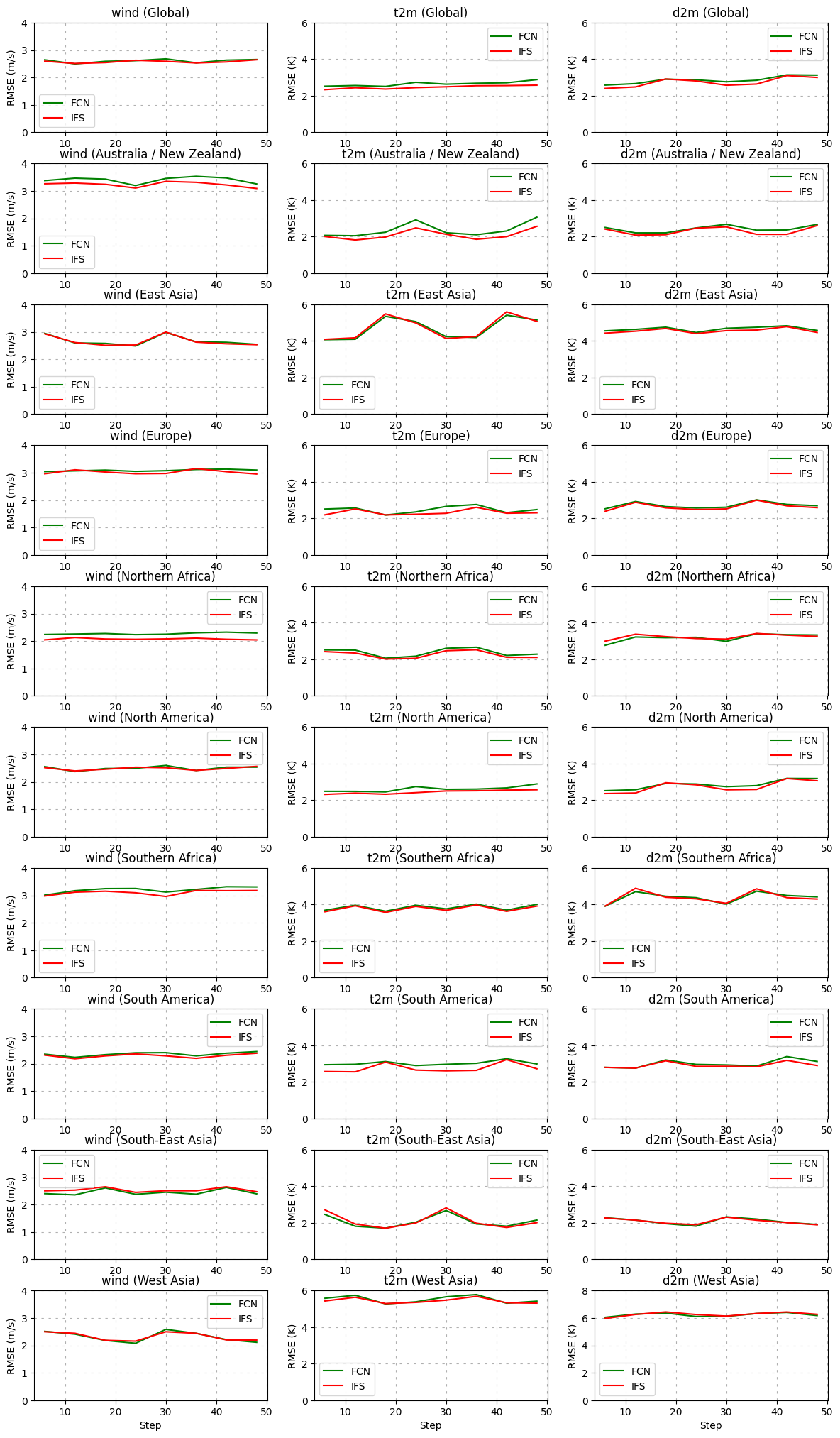}
    \caption{RMSE for different variables, comparing FCN and IFS against observations for different regions of the world. Each row is one region, and the columns represent the RMSE for wind speed, temperature, and dewpoint respectively.}
    \label{fig:madiseval}
\end{figure}

Figure~\ref{fig:madiseval} shows the performance of FCN and IFS for all variables different regions, for a 48 hour forecast. It is immediately evident that both FCN and IFS perform nearly identically in all regions and for all lead times, which contradicts our expectation that FCN should outperform IFS, given how well it does when tested against ERA5. This result underscores the need for testing DDWPs in realistic settings as they are deployed in operational forecasts.

%\vspace{-2mm}
\section{Discussion}
%\vspace{-1mm}
DDWPs have made significant improvements in a short time, and show great promise for use in operational weather forecasting. So far, DDWPs have been evaluated using reanalysis data as ground truth, which effectively tests DDWPs in a high-quality simulation. While valuable, such validation is incomplete and more thorough evaluation is needed under settings that replicate their operational use. While by no means exhaustive, the use of in-situ observations from MADIS2020 provides one such way to evaluate DDWPs on a meaningful task, with the benefit that they can be used to fairly compare all DDWPs and NWPs. Compared to ERA5 data, real-world observations better represent the conditions experienced by end-users of weather forecasts, making such an evaluation benchmark particularly relevant when deploying DDWPs in operational forecasts. Indeed, when comparing one well-known DDWP model (FourCastNet) against a baseline NWP (the IFS model), we see that any performance advantage shown when testing against ERA5 disappears when testing against real-world observations.\par

While further work is needed to identify why DDWPs may not perform well against real-world data, one reason for the difference in performance comes from the use of less accurate start states in operational settings. This may be addressed by fine-tuning DDWPs on operational start states once they are sufficiently trained on ERA5 data. The ERA5 dataset itself can be further examined and uesd more effectively in training DDWPs. The uncertainty of ERA5 reanalysis varies by time and location and tends to be higher in the global South, which can in turn influence the quality of any models trained on ERA5 data. One effect of such variation in data quality is seen in the regional variation of model performance, and ML practitioners must take care to ensure their evaluations do not focus purely on the data-rich parts of the world.  The ensemble "spread" available in ERA5 conveys a specific type of uncertainty estimate, and may be used to improve the training of DDWPs by adjusting how much to weight data from different locations and times. The uncertainty is expected to also be higher the farther back in time we go (due to sparser observations), and accounting for uncertainty may bias DDWPs towards recent data and help reduce any training-inference skew caused by weather patterns changing due to climate change.\par

As weather forecasts are often consumed at specific locations of high interest like airports, interpolation from a grid to a specific latitude/longitude is a necessary step. Bilinear interpolation is a common technique, but ML could be applied to perform data-driven interpolation, taking into account sub-grid features like topography, terrain, land use, etc., for which in-situ observations are particularly useful as training and evaluation data. Improved interpolation techniques could once again exploit observations from data-rich regions for the benefit of data-poor regions. Such techniques can be compared against techniques like Model Output Statistics (MOS), used to provide site-specific guidance derived from the gridded GFS forecast issued by NOAA (\cite{glahn1972use}). Where a long history of observations is available, data-driven debiasing can be used to remove systemic biases present in weather models. The definition and use of appropriate metrics is necessary to build trust amongst stakeholders and properly evaluate the quality of DDWPs. While deterministic metrics like RMSE are convenient to report, they are not appropriate for all variables (e.g. precipitation) and for longer lead times. For longer lead times, it may be more appropriate to use probabilistic metrics as well as NWP ensembles as a more relevant baseline.\par
The recent activity and progress in researching ML-based forecasting leaves little doubt that data-driven weather prediction models will have a meaningful role to play in operational forecasting. While much of the recent work has been focused on learning an NWP approximation in an ERA5 simulation, imagining a DDWP in an operational setting raises many more interesting questions (and avenues for research) around their actual use. Our work should be seen as the start of a conversation around answering such questions to realize the full potential of ML-based weather forecasting.

\begin{ack}
The authors would like to thank the European Center for Medium Weather Forecasting (ECMWF) for making available the ERA5 dataset and the TIGGE archived NWP forecasts, used as IFS in the study. Likewise, we would like to thank NOAA for making publicly available such a large corpus of in-situ observations from the MADIS program.
\end{ack}

\newpage

\bibliographystyle{apalike}
\bibliography{refs}

\newpage

\section{Appendix for \texttt{MADIS2020} Dataset}

In this section we will provide a brief overview of the dataset directory, code snippets to access and manipulate the data and provide links for open-source implementations for the model results with the goal to improve reproducibility.

\section*{Dataset directory structure}

The \texttt{\texttt{MADIS2020}} dataset contains observations from two separate networks, namely

\begin{enumerate}
    \item \textbf{METAR}: Primarily a format for reporting weather information that is predominantly used in aviation and provided by airports. 
    \item \textbf{MESONET}: These are data obtained from private weather stations that measure and collect data that are important to the purposes of providing a comprehensive meteorological understanding.
\end{enumerate}

The directory follows the following structure

...
\dirtree{%
.1 \texttt{madis2020}.
.2 \texttt{month}.
.3 metar.
.4 zarr datafile.
.3 mesonet.
.4 zarr datafile.
}

The MADIS observation data is stored in \texttt{zarr} files, which can be opened using the open-source Xarray library \cite{hoyer2017xarray}. The \texttt{month} range is between 1 (for January) to 12 (for December). Each \texttt{zarr} data file comprises observation for a single hour, therefore totaling 24 files for the same date. The observations for a specific hour \texttt{YYYY-MM-DD HH:00} are stored in the file path \texttt{\texttt{MADIS2020}}/\texttt{month}/\texttt{(mesonet|metar)}/\texttt{YYYYMMDD\_HH00.zarr}.

Each \texttt{zarr} file has the variables (as shown in Table \ref{var-table}) with quality control flags (more on NOAA website \footnote{\url{https://madis.ncep.noaa.gov/madis_sfc_qc_notes.shtml}}), however it is to be noted that the \texttt{accumulated} precipitation variables are only available for \texttt{METAR}.

Lastly, the dataset contains two variables \texttt{latitude} and \texttt{longitude} that specify where the observation was taken.

\section*{Example codes}

In this section we will share some basic code snippets to access and manipulate the data for evaluation, assuming the data is in a standard format such as NetCDF or Zarr.
\subsection*{Code to access data}
\begin{lstlisting}[language=Python]
# Import python library
import xarray as xr

# Load file path
FILE_PATH  = '<DIRECTORY_PATH>/madis2020/<month>/<metar|mesonet>/YYYYMMDD_HH00.zarr'

# Access data from file
hourly_data = xr.open_zarr(FILE_PATH)

# Shows all variables including quality flags, latitude, and longitude
weather_vars = hourly_data.data_vars
\end{lstlisting}

\subsection*{Code to apply quality flags}

The details regarding the \texttt{QC} flags have been discussed extensively in the main paper. Here we provide a snippet of code examples to apply some flags. Further details on the \texttt{QC} levels are available on the NOAA website \footnote{\url{https://madis.ncep.noaa.gov/madis_sfc_qc_notes.shtml}}.

\begin{lstlisting}[language=Python]
# apply relevant quality flag
valid_observation = ((hourly_data.windSpeedDD == b'S') + (hourly_data.windSpeedDD == b'V'))

# code to ignore outliers
validobs = valid_observation & (hourly_data.windSpeed < 50)

# extract data
speed_obs = hourly_data.windSpeed[validobs].load()
lat_obs = hourly_data.latitude[validobs].load()

# add longitude conversion if required in the following step
lon_obs = (hourly_data.longitude[validobs].load()
\end{lstlisting}

This returns the arrays for wind observations (\texttt{speed\_obs}) at relevant locations (\texttt{lat\_obs} and \texttt{lon\_obs}). The model prediction at relevant locations can then be used to compare against obervations.

\subsection*{Comparing observations to model outputs}

Given that ground observations are location-specific, the model outputs which are often times in a gridded format, are required to interpolate from the neighboring nodes to the relevant latitude/longitude in question. The reader can use any standard package, for example, Xarray's data array interpolation \footnote{\url{https://docs.xarray.dev/en/stable/generated/xarray.DataArray.interp.html}} for this purpose. We present a code snippet on how to evaluate one hour of ERA5 data against MADIS observations, noting that similar procedures can be used for any modeled data.

\begin{lstlisting}[language=Python]
# Import xarray library
import xarray as xr
import numpy as np

# Step 1: Load the hourly data, and filter for valid temperature observations.
FILE_PATH  = '<DIRECTORY_PATH>/madis2020/<month>/<metar|mesonet>/YYYYMMDD_HH00.zarr'
madis_data = xr.open_zarr(FILE_PATH)
valid_observation = ((madis_data.temperatureDD == b'S') + 
                     (madis_data.temperatureDD == b'V'))
valid_lats = madis_data.latitude[valid_observation]
valid_lons = madis_data.longitude[valid_observation]
valid_t2m = madis_data.temperature[valid_observation] 

# Step 2: Load the ERA5 data for temperature, for the same date and time.
# The ERA5 data can be downloaded directly from ECMWF's CDS API,
# and a reference script to download data is available from
# WeatherBench:
# https://github.com/pangeo-data/WeatherBench/blob/master/src/download.py
# In this example, the NetCDF file is expected to contain data for 24
# hours of January 1, 2020. Such data can be downloaded by specifying
# the following flags to the download script from WeatherBench:
# --variable=t2m --level_type=single --years 2020 --month 1 --day 1
# Note that this would download data at a 0.703125 degree resolution.
ERA5_PATH = '<ERA5_DIRECTORY>/t2m_20200101.nc'
era5_data = xr.open_dataset(ERA5_PATH)

# Step 3: Interpolate the ERA5 data for hour 0 to the required latitude/longitudes.
# Note that the actual resolution of the data is encoded in the NetCDF file,
# and we can exploit xarray's functions to perform bilinear interpolation
# for all desired points.
era5_pred = era5_data.t2m[0].interp(latitude=xr.DataArray(valid_lats),
                                    longitude=xr.DataArray(valid_lons))

# Step 4: Compute the error between the observed temperature, and
# the ERA5 modeled temperature.
error = (valid_t2m.to_numpy().flatten()
         - era5_pred.to_numpy().flatten())
mse = np.mean(error**2)
\end{lstlisting}

\section*{Bias in the Dataset}

\begin{figure}[htb]
    \centering
    \includegraphics[scale=0.6]{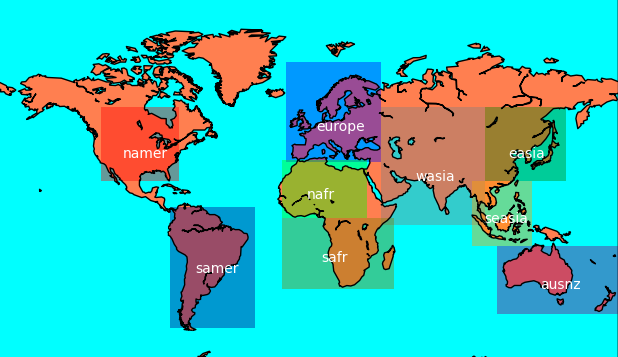}
    \caption{Geographical extent of various regions used in evaluations.}
    \label{figa1}
\end{figure}

We conducted a study to verify any location-specific biase in the dataset and divided the globe into specific regions, graphically shown in Figure \ref{figa1} and the latitude/longitude range is shown in Table \ref{geo-table}.

\begin{table}[htb]
  \caption{The latitude and longitude extents of the various geographies used in the study}
  \label{geo-table}
  \centering
  \begin{tabular}{lll}
    \toprule
    Geographic Region & Longitude range &  Latitude range \\
    \midrule
    \texttt{global} & $\mathtt{-180}$\textdegree{} to $\mathtt{180}$\textdegree{} & $\mathtt{-90}$\textdegree{} to $\mathtt{90}$\textdegree{} \\
    \texttt{namer} (North America)&  $\mathtt{-120}$\textdegree{} to $\mathtt{-75}$\textdegree{} &  $\mathtt{25}$\textdegree{} to $\mathtt{60}$\textdegree{} \\
    \texttt{europe} (Europe) & $\mathtt{-12.5}$\textdegree{} to $\mathtt{42.5}$\textdegree{} &  $\mathtt{35}$\textdegree{} to $\mathtt{75}$\textdegree{} \\
    \texttt{wasia} (West Asia) &  $\mathtt{42.5}$\textdegree{} to $\mathtt{102.5}$\textdegree{} &  $\mathtt{0}$\textdegree{} to $\mathtt{60}$\textdegree{} \\
    \texttt{easia} (East Asia)&  $\mathtt{102.5}$\textdegree{} to $\mathtt{150}$\textdegree{} &  $\mathtt{25}$\textdegree{} to $\mathtt{60}$\textdegree{} \\
    \texttt{seasia} (South-East Asia) &  $\mathtt{95}$\textdegree{} to $\mathtt{130}$\textdegree{} &  $\mathtt{-12.5}$\textdegree{} to $\mathtt{25}$\textdegree{} \\
    \texttt{ausnz} (Aus. \& New Zealand)&  $\mathtt{110}$\textdegree{} to $\mathtt{180}$\textdegree{} &  $\mathtt{-48}$\textdegree{} to $\mathtt{-12.5}$\textdegree{} \\
    \texttt{samer} (South America) &  $\mathtt{-80}$\textdegree{} to $\mathtt{-31}$\textdegree{} &  $\mathtt{-54}$\textdegree{} to $\mathtt{10}$\textdegree{} \\
    \texttt{nafr} (Northern Africa) &  $\mathtt{-15}$\textdegree{} to $\mathtt{34}$\textdegree{} &  $\mathtt{4}$\textdegree{} to $\mathtt{36}$\textdegree{} \\
    \texttt{safr} (Southern Africa) &  $\mathtt{-15}$\textdegree{} to $\mathtt{50}$\textdegree{} &  $\mathtt{-36}$\textdegree{} to $\mathtt{4}$\textdegree{} \\

    \bottomrule
  \end{tabular}
\end{table}

In Figure \ref{fig:obscount} we plot the data distribution across various geographies and it appears that the global south and the APAC region suffer disproportionately with comparably fewer high-quality ground observations. This trend exacerbates for variables such as precipitation which while only available for \texttt{METAR}, has a huge discrepancy for non-North American geographies.

\section*{Benchmarking}

We evaluate three sources of weather data against real-world observations, - the ERA5 reanalyses product from ECMWF, FourCastNet data driven model and IFS Numerical Weather Prediction model made available by ECMWF. Full details of the FourCastNet model implementation are discussed in \cite{pathak2022fourcastnet} and the original model implementation is publicly available \footnote{\url{https://github.com/NVlabs/FourCastNet}}. For the IFS forecast, the TIGGE archived NWP forecasts were used \footnote{\url{https://www.ecmwf.int/en/research/projects/tigge}}. The ERA5 dataset \footnote{\url{https://www.ecmwf.int/en/forecasts/dataset/ecmwf-reanalysis-v5}} is also made publicly available by ECMWF. 

\begin{figure}[!htb]
    \centering
    \includegraphics[scale=0.4]{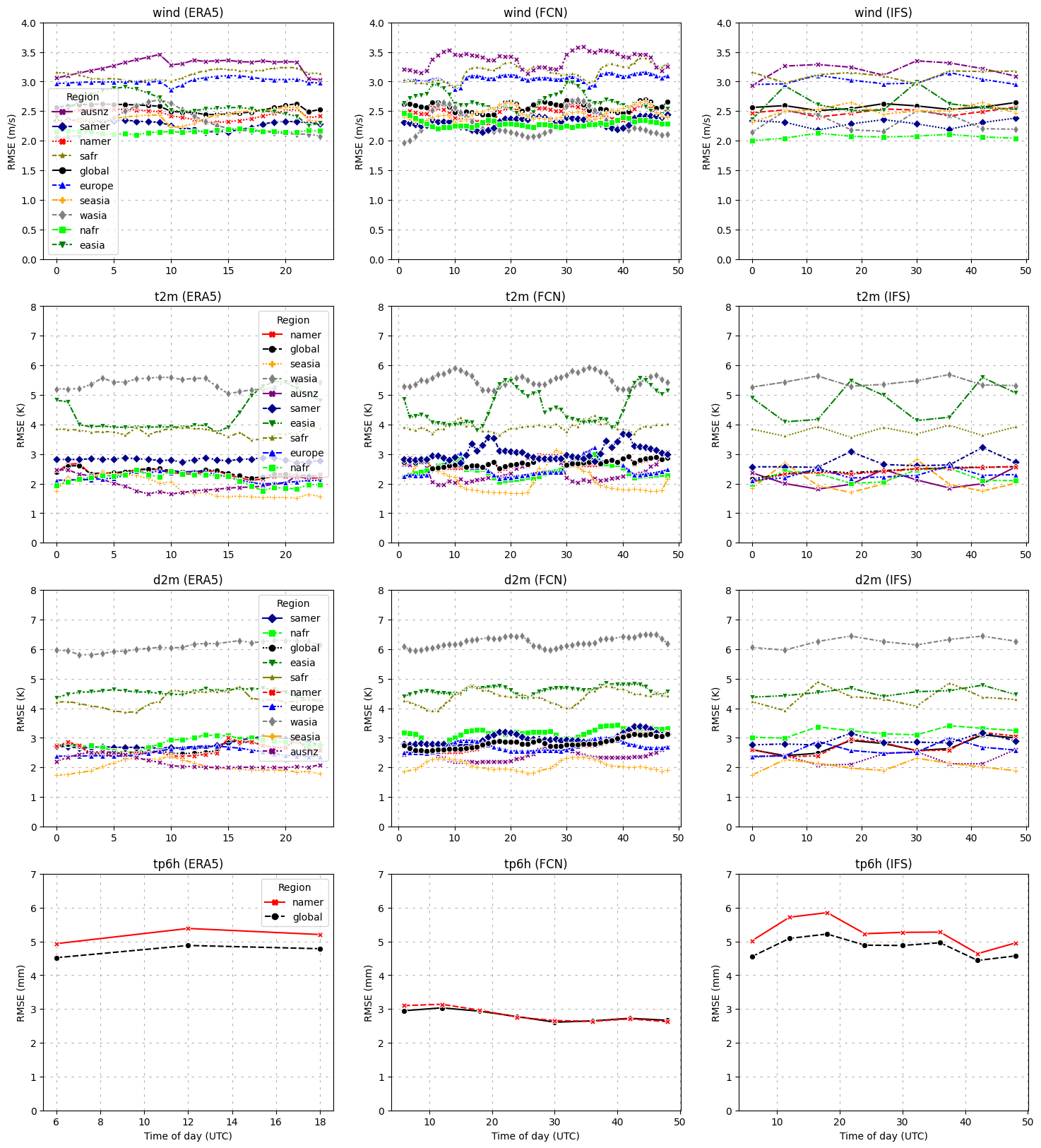}
    \caption{RMSE of ERA5, FCN, and IFS against observations for wind speeds, temperature, dewpoint, and 6-hourly precipitation accumulation. Each column represents a data source, and each row represents a variable. The x-axis represents the time of day in UTC, and the y-axis represents the RMSE in appropriate units. Note the different forecast depths for ERA5 vs. FCN and IFS.}
    \label{fig:varplots}
\end{figure}

The all-weather verification in Figure \ref{fig:varplots} shows a close comparison between the three sources of data with signs of uncorrelated errors between the data-driven (FCN) and numerical weather prediction (IFS) model.

%\bibliographystyle{apalike}
%\bibliography{refs}

%%%%%%%%%%%%%%%%%%%%%%%%%%%%%%%%%%%%%%%%%%%%%%%%%%%%%%%%%%%%
\end{document}